\documentclass[acmtog, nonacm]{acmart}
\pdfoutput=1

\usepackage{hyperref}
\usepackage{booktabs} % For formal tables
\usepackage{microtype}
\usepackage{amsmath}
\usepackage{overpic}
\usepackage[ruled]{algorithm2e} % For algorithms

\SetAlFnt{\small}
\SetAlCapFnt{\small}
\SetAlCapNameFnt{\small}
\SetAlCapHSkip{0pt}

\DeclareMathOperator*{\argmin}{arg\,min}

\setcitestyle{square}

\author{Meng Zhang}
\affiliation{ \institution{University College London} 
\country{United Kingdom}} 

\author{Duygu Ceylan}
\affiliation{ \institution{Adobe Research} \country{United Kingdom}} 

\author{Tuanfeng Wang}
\affiliation{ \institution{Adobe Research} \country{United Kingdom}} 

\author{Niloy J. Mitra}
\affiliation{ \institution{University College London and Adobe Research} \country{United Kingdom}} 

\renewcommand{\shortauthors}{Zhang et al.}
\renewcommand\footnotetextcopyrightpermission[1]{}
\begin{document}

% Title. 
\title{Dynamic Neural Garments}
\thispagestyle{empty}  

% abstract
\fancyfoot{}
\begin{abstract}
A vital task of the wider digital human effort is the creation of realistic garments on digital avatars, both in the form of characteristic fold patterns and wrinkles in static frames as well as richness of garment dynamics under avatars' motion. 
Existing workflow of modeling, simulation, and rendering closely replicates the physics behind real garments, but is tedious and requires repeating most of the workflow under changes to characters' motion, camera angle, or garment resizing. 
Although data-driven solutions exist, they either focus on static scenarios or only handle dynamics of tight garments. 
We present a solution that, at test time, takes in body joint motion to directly produce realistic dynamic garment image sequences. 
Specifically, given the target joint motion sequence of an avatar, we 
propose \emph{dynamic neural garments} to jointly simulate and render plausible dynamic garment appearance from an unseen viewpoint. Technically, our solution generates a coarse garment proxy sequence, learns deep dynamic features attached to this template, and neurally renders the features to produce appearance changes such as folds, wrinkles, and silhouettes.
We demonstrate generalization behavior to both unseen motion and unseen camera views. Further, our network can be fine-tuned to adopt to new body shape and/or background images. 
We also provide comparisons against existing neural rendering and image sequence translation approaches, and report clear quantitative improvements. 
\end{abstract} 

%CCS

%keywords

\begin{teaserfigure}
  \includegraphics[width=\textwidth]{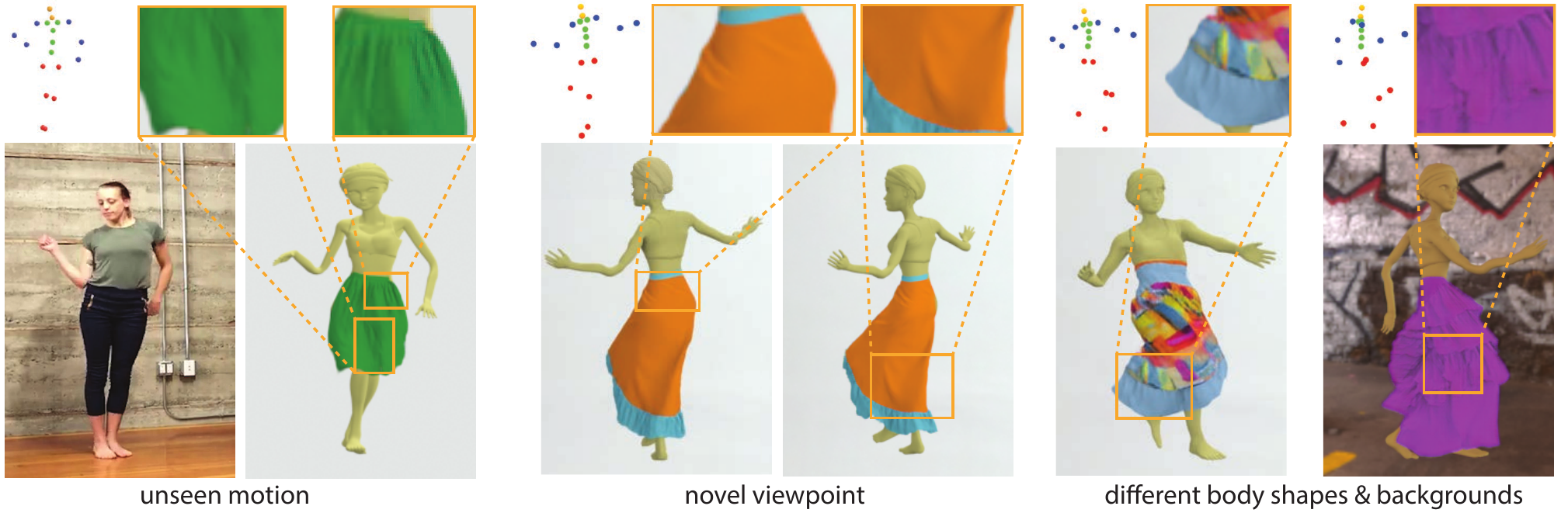} 
  \caption{Given a 3D body motion sequence (potentially extracted from real capture), we present \emph{Dynamic Neural Garments} that jointly learns to simulate and synthesize the dynamic appearance of a target garment and harmonizes it with the background rendering of an undressed 3D character. Once trained, our network synthesizes the appearance of the garment with different motions and viewpoints \emph{without} running physically-based simulation. It can also be fine-tuned to generalize to new body shapes and background images demonstrating different illumination conditions.} 
  \label{fig:teaser}
\end{teaserfigure}

\maketitle
\thispagestyle{empty}

\section{Introduction}

\begin{quote}
    {\em A great dress can make you remember what is beautiful about life.} \hfill {Rachel Roy}
\end{quote}

% demand for realistic garments
In the real world, we observe a wide range of garments on humans. Such garments display characteristic details (e.g., crease and folds) arising from a variety of factors, including their stitching layout, underlying materials, or printed patterns on the base fabrics. Naturally, there is a strong motivation, both for games and VR applications, to similarly `clad' virtual avatars in realistic garments.

A popular workflow, enabled by commercial software (e.g., Marvellous Designer), is to mimic the real world in every stage: geometrically model a garment mesh using real-world stitching patterns for guidance, anticipate mesh dynamics using a physically-based simulator under the action of the body movements, and finally render the resultant dynamic mesh using available texture information. Achieving realistic results using this workflow is tedious and expensive due to the high level of precision required across the modeling, simulation, and rendering stages. Further, much of the process has to be repeated when any of the garment details, body motion, or viewing camera location need to be updated. 
A modified workflow encourages using a coarse garment model that is physically-simulated and high-resolution details added in a postprocessing stage using baked-in texture maps. While the approach is efficient and robust, the skinned garments appear stiff as the baked-in details do not move realistically on loose garments.

%neural way
Recently, data-driven approaches have been proposed to replace parts of the above process. For example, hallucinating geometric details on simulation of coarse garment templates~\cite{lahner2018deepwrinkles,Zhang2020}, extending parametric human body models with per-vertex displacements to capture the deformation of tight garments \cite{alldieck2019learning}, or directly translating image-space body joint locations to final rendered images to recreate person-specific dance sequences~\cite{chan2019dance}. 
Another intriguing alternative to produce realistic renderings is to train a network to produce detailed high-quality (rendered) images that can be directly supervised with multi-view image information~\cite{thies2019deferred,mildenhall2020nerf}. %that avoids the requirement for explicit geometric modeling~\cite{mildenhall2020nerf}. \meng{(because "2019deferred" needs a explicit geometric modeling)} 
Such neural rendering approaches~\cite{RenderNet2018,thies2019deferred}, however, are mostly restricted to static objects and cannot handle intricate dynamics like the movement of garments under the action of the underlying body movements. An exception and particularly relevant to ours is the work of Chen et al.~\shortcite{chan2019dance} that handles dynamic sequences with tight garments on actors. The method, however, does not generalize to produce dynamics of loose fabrics, as in our focus (see Section~\ref{subsec:results} for comparison).

%desirable properties
In this work, we consider the problem of estimating detailed garment dynamics and generating realistic image sequences with plausible details driven by the given 3D (human) body motion information. %\meng{How about: "In this work, we consider the problem as to achieve the detailed garment construction and vivid moving variance in the rendering stage driven by the given 3D(human) body motion information. ..." }
A good solution should (i)~produce realistic image-level garment details that are consistent across frames and camera views; (ii)~produce plausible garment motion in response to any input body motion; %(iii)~avoid requiring modeling garments in full geometric details or relying on precise simulation for training; 
(iii)~be flexible enough to be trained using multiview target appearance supervision; (iv)~independent of physically-based simulation at test time; and (v)~generalize across motion and view variations.

\begin{figure*}[t]
  \includegraphics[width=\textwidth]{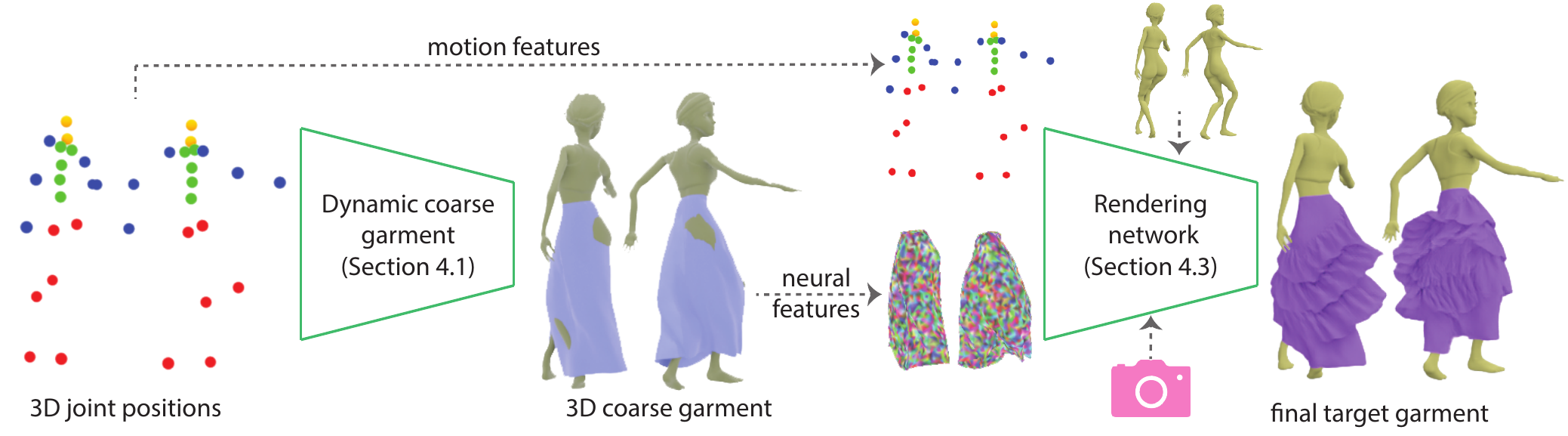}
  \caption{Starting from a body motion sequence, we first synthesize how a coarse garment proxy would deform in 3D. We then learn deep dynamic features on the coarse template to synthesize the final appearance of the target garment from a desired viewpoint. Note that the target garment is different from the coarse template geometrically and hence both the overall outline of the garment and the dynamic appearance, i.e., folds, silhouettes, and wrinkles, are significantly different.}
  \label{fig:overview}
\end{figure*}

%our algorithm
To this end, starting from only a sequence of body joint information, we predict a  coarse proxy template sequence over frames and learn dynamic deep features attached to this coarse template. The learned features capture both the geometric differences between the coarse and target garment (e.g., lace structure in Figure~\ref{fig:overview}) and the dynamic appearance changes (e.g., appearance/disappearance of folds under body motion). The features are dynamic in two ways: first, the overall movement is captured by their geometric displacement across frames (i.e., coarse template motion), and second, the learned neural features are concatenated with body joint information before being interpolated and rendered.  By projecting and rendering these features from input camera views, we train the whole process using image information while using intermediate coarse simulation as a proxy loss (i.e., our current realisation is not end-to-end trained). Finally, in order to ensure temporal smoothness, we propose an encoder structure that explicitly models the correlation between the temporal latent codes as well as a temporal discriminator to ensure synthesized consecutive frames are temporally coherent.

We evaluate our algorithm on a range of examples with varying body motion and target garment styles. We demonstrate that our network generalizes over a range of viewpoints and body motion. Further, the approach handles changes to body sizes and illumination conditions (provided via a background image) with fine-tuning. We compare with strong baselines including image translation networks~\cite{isola2017image,chan2019dance} and deferred neural rendering~\cite{thies2019deferred} to demonstrate clear improvements. Compared to the strongest baseline, our method achieves an improvement of $70\%$ and $46\%$ in the image based~\cite{heusel2017gans} and video based~\cite{wang2018vid2vid} FID scores, respectively. Our is efficient running at $16$ fps, while traditional physically-based simulation approaches can only achieve $0.5$ fps on average for similar visual quality. %This brings a great potential for a real-time system in the future.

In summary, our key contributions are: 
\begin{itemize}
    \item \emph{motion driven rendering} where the input 3D body motion sequence is used to generate realistic pixel-level frame renderings of complex target garments; 
    \item \emph{dynamic deferred neural rendering} based on an intermediate coarse proxy to synthesize varying motion appearance with the target colors and textures; 
    %\item \emph{temporally coherent encoder} to implicitly defer the variance of the target motion appearance by the correlation computation between the temporal latent codes;
    %\item \duygu{everybody dance now also uses a similar temporal discriminator, we should not claim it as a contribution?} \emph{temporal-and-spatial discriminator} to constrain the context consistency and continuity in both time and space; and 
    \item \emph{generalization} across views and motions to effectively avoid explicitly modeling, rendering, and simulating garment sequences on digital avatars for every new motion sequence. 
\end{itemize}

\section{Related Work}
\textbf{Garment authoring.}
Starting from an initial, possibly coarse garment, several methods have been proposed for intuitive editing of the shape and appearance of the garment. Umetani et al.~\cite{Umetani:2011} propose an interactive system that enables to edit both 2D sewing patterns and the 3D draped garments. In a similar fashion, Bartle et
al.~\cite{Bartle:2016} present a method to map 3D edits to plausible
2D garment patterns. Several methods focus on specific type of edits such as resizing~\cite{meng2012flexible} or stylizing tight-fitting garments~\cite{kwok2016styling}. The \emph{SecondSkin} system proposed by De Paoli et al~\shortcite{Paoli2015} focuses on modeling layered accessories and garments directly on a 3D character. More recently, Li et al.~\shortcite{Li:2018:FoldSketch} has presented an intuitive system to add folds and pleas to a garment. All of these approaches work on a static garment, while we focus on generating dynamics of a given garment. Hence our work is complimentary, our system can be used to learn the dynamic appearance of a garment authored by the previous methods.

\textbf{Modeling garment dynamics.} 
Physically based simulation provides an accurate way to model garment dynamics~\cite{Ko2005,Nealen2006,narain2012adaptive,liang2019differentiable,yu2019simulcap,tang2018cloth}. However, as the complexity of the garments increases, computational cost and stability issues become major sources of concern. Hence, several methods have been proposed to approximate the costly physically based simulation process. One line of work utilizes constraint-based optimization methods to compute a high resolution mesh from a coarse one~\cite{Muller2010,Rohmer2010,Gillette2015}. Other popular approaches include utilizing data-driven methods to generate high resolution garment meshes by learning a mapping from a coarse garment mesh to fine-scale displacements~\cite{Feng2010,zurdo2012animating}, utilizing learned upsampling operators~\cite{Kavan2011}, interpolating and blending example samples in the database~\cite{Wang2010,xu2014sensitivity}, or learning a subspace model of garment deformations~\cite{guan2012drape,hahn2014subspace}.

In recent years, researchers have explored the use of deep learning methods to learn how garments deform under body motion. In case of tight clothing (i.e., uv coordinates on body closely match uv coordinates of garments), representing garments as per-vertex displacements with respect to the naked body is a common and efficient practice~\cite{alldieck2019learning,bhatnagar2019multi,pons2017clothcap,jin2018pixel,ma2019learning}. However, this approach falls short in modeling loose garments. One approach to address this limitation is to utilize recent implicit based representations to recover animatable 3D characters given input images~\cite{huang_arch_2020,Yang2021}. However, such approaches do not model the garments separately and are limited in terms of modeling the complex garment dynamics. Hence, several works assume the template garment to be known at rest pose and learn to construct the deformed garment shape under different body poses. While some methods utilize subspace techniques to operate on a reduced deformation space~\cite{yang2018analyzing, wang2018learning, holden2019subspace}, others directly learn to predict the deformed garment shape~\cite{gundogdu2019garnet,santesteban2019learning,patel2020virtual}. Finally, some recent work augment a low-resolution of a garment normal map with plausible wrinkles~\cite{lahner2018deepwrinkles,Zhang2020}.

In our method, we use a similar learning based approach to predict the deformed shape of a coarse garment under body motion~\cite{wang2019learning}. Unlike previous methods, however, our method adopts a neural rendering approach to synthesize garment deformations with rich details under different styles and viewpoints. The coarse template required by our method is not only a downsampled version of the desired garment but also is free of certain geometric details such as laces or multi-layer components which are hard to model and simulate. We show that starting from the same coarse garment deformation, we can plausibly synthesize different high resolution garment styles under various body motion~(Section \ref{sec:experiments}).

\textbf{Neural rendering.}
We are recently witnessing an exciting breakthrough in the field of \emph{neural rendering} where deep neural features are learned for controllable image synthesis including viewpoint change and modeling deformations~(see \cite{Tewari2020} for a survey). Neural features have been learned on various representations including voxel grids~\cite{sitzmann2019deepvoxels,mildenhall2020nerf,liu2020neural}, 3D meshes~\cite{thies2019deferred}, point clouds~\cite{aliev2019neural}, and multi-layer images~\cite{Lu2020}. Most of these approaches, however, focus on modeling viewpoint changes or illumination changes for static scenes or objects. Most recently, these methods have been extended to handle non-rigidly deforming~\cite{park2020nerfies,tretschk2020nonrigid} and dynamic scenes~\cite{pumarola2020d,LiDynamic2020}.

In the context of humans, neural rendering has been utilized to synthesize faces under different head pose and expression~\cite{Lombardi:2019,wang2020learning}, render humans under different viewpoint and illumination~\cite{Meka:2020}, and for reposing~\cite{Shysheya_2019_CVPR,Sarkar2020}. Most of these approaches assume humans wear tight clothing and do not explicitly focus on modeling the dynamics of loose garments. Our work aims to fill this void by proposing a novel neural rendering pipeline to render plausible dynamics for loose garments that generalizes across motion changes and view variations.

\textbf{Image-to-image translation.}
Image-to-image translation methods have shown incredible success recently in terms of translating images from a particular domain to another~\cite{liao2017visual,fivser2016stylit,huang2017arbitrary,isola2017image}. These methods have also been extended to the temporal domain to synthesize videos from a guiding signal such as segmentation masks~\cite{wang2018vid2vid,mallya2020world}. Inspired by the success of such methods, several works have explored these translation methods to animate various objects.

Many video-based translation methods utilize an intermediate representation based on keypoints~\cite{Siarohin_2019_CVPR,Siarohin_2019_NeurIPS,Minderer2019} and transfer the motion of a source video to a different object. In the context of animating humans and virtual try-on, keypoint representations have been replaced with 2D pose~\cite{Aberman2019,chan2019dance,Dong_2019_ICCV}, human body part segmentation~\cite{Zhou_2019_ICCV}, and dense correspondences obtained from a 3D body mesh~\cite{Liu_2019_ICCV,Zablotskaia2019}. The work of Chan and colleagues~\shortcite{chan2019dance} is particularly relevant for our focus and we provide an explicit comparison in Section~\ref{sec:experiments}.
Most of these methods model humans in tight clothing, however, and do not focus on garment dynamics. In contrast, our method learns dynamic deep garment features together with a corresponding neural renderer to synthesize high quality deformations for loose garments.

\section{Overview}
%\begin{figure}[t]
%  \includegraphics[width=\columnwidth]{fig/overView_1.png}
%  \caption{overview}
%  \label{fig:overview1}
%\end{figure}
Given the rendering of a character with a target garment $V$ under certain motion sequences at training time, our method learns to synthesize the image space appearance of the garment over the character given a new (unseen) body motion sequence $M_t$ (where $t$ refers to the frame number in the sequence) and user-specified camera parameters $p$. 

Directly generating highly dynamic garment appearance, especially loose garments, only from the body joint motion is challenging due to the highly nonlinear nature of garment dynamics. Hence, we first learn 3D dynamics of a coarse garment proxy, $V^c$, as an intermediate representation. Then, at the core of our approach, we propose a neural rendering technique that learns dynamic neural features over the coarse garment template along with a neural renderer that interprets these features to synthesize the final appearance of the target garment from the desired viewpoint. As shown in Fig.~\ref{fig:overview}, given the input character motion $M_t$ represented as the trajectory of body joints, we first train a network to drive the pre-defined coarse garment in 3D resulting in $V^c_t$ (Section~\ref{subsec:CoarseTemplate}). The coarse template is not only a downsampled version of the target garment but also lacks detailed geometric features such as laces or multi-layer components. Hence, the \textit{same} coarse template can be shared by different target garments of the same type (e.g., a coarse garment in the form of a plain skirt can drive both a pleated and a laced skirt). In other words, the dynamics of the coarse garment can be shared across different garment styles. 

Inspired by the recent work of Thies et al.~\shortcite{thies2019deferred}, we learn a $d$-dimensional neural texture $F$ for the coarse garment and generate a neural feature image $F^p_t$ from the desired viewpoint by sampling the texture based on the predefined UV coordinates of $V^c$. In order to capture the dynamics of the garment, we also define a motion descriptor and generate another feature image $S^p_t$ that encodes the body motion. Our dynamic neural rendering network $G$ synthesizes the final garment appearance conditioned on the two neural feature images and a sequence of rendered background images $\{B^p_t\}$ which provide the appearance of the character body without the garment (Section~\ref{subsec:RenderingNet}). 

In order to ensure the generated garment appearance blends well with varying illumination and character body rendering styles, $G$ uses a multi-layer perceptron (MLP) based decoder to further harmonize the synthesized garment with the background images. Our neural rendering component is trained end-to-end to learn both the neural texture $F$ and the parameters of the rendering network $G$ based on the objective function:

\begin{equation*}
    F^\star,G^\star := \argmin_{F,G}\sum_{t,p}\mathcal{L}(G(F^p_t,S^p_t,B^p_t),I_t^p)
\end{equation*}
where $\mathcal{L}$ is a suitable training loss computed on the generated image and the ground truth image $I_t^p$. We also adopt an adversarial setup by applying a spatial-and-temporal discriminator to improve the accuracy and consistency of the synthesised output. Next, we describe the individual steps in detail.

\section{Algorithm}

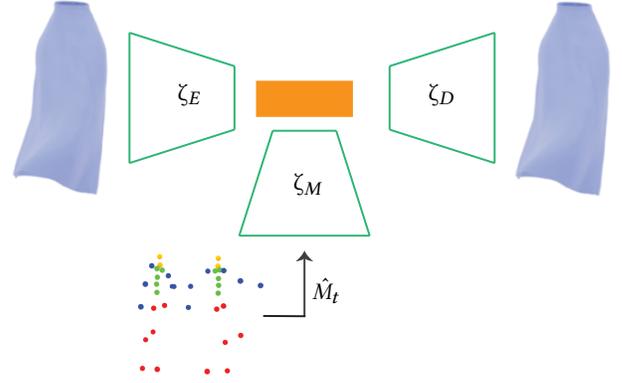
\begin{figure}[!t]
  \begin{overpic}[width=\columnwidth]{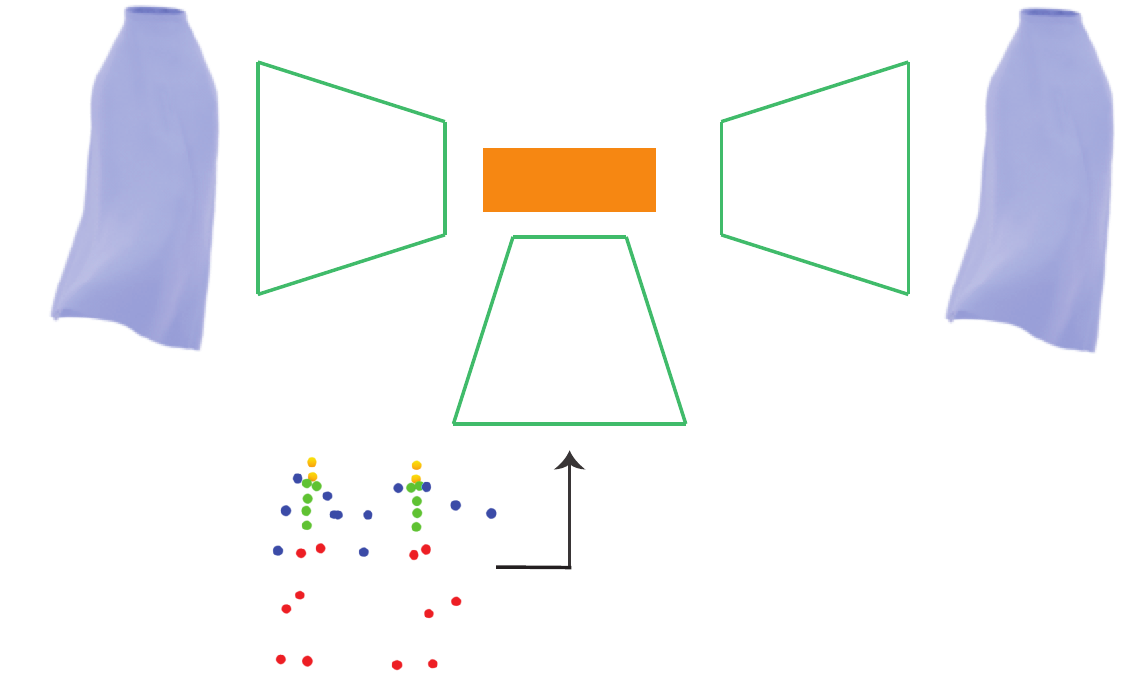}
    \put(30,44){$\upzeta_E$}
    \put(69,44){$\upzeta_D$}
    \put(48,30){$\upzeta_M$}
    \put(51,13){$\hat{M}_t$}
  \end{overpic}
  \caption{Given 3d body joint positions of a motion sequence, we train an encoder-decoder type of network to map the motion information to the learned latent space of a coarse garment template and decode its 3D deformed shape.}
  \label{fig:coarse}
\end{figure}

\subsection{Dynamic Coarse Garment}\label{subsec:CoarseTemplate}
%\subsection{Coarse neural template}\label{subsec:CoarseTemplate}
%In the DRAPE~\cite{guan2012drape}, they deform the template garment to fit any new body and pose. While, the deformation has the well-defined vertex-wise corresponding, since the template and the target shares the same garment style and geometric topology. Our coarse template aims at providing the  implicit common features for the garment animation to guide the neural features and rendering network to learn the target-style specified appearance. Our template plays a role a bit similar to the template face in the task of face animation~\cite{cao2014displaced}. However, our target styled motion rendering is generated by our rendering network and neural features rather than image wrapping combining with the detail enhancement. Our rendering task sets a goal of using one coarse simple-styled skirt to drive multiple complex designed skirt animation with fine-grained wrinkles and dramatic skirt dancing which are actually unseen in the template animation.
Given a desired body motion sequence $M_t$, our method first synthesizes the dynamics of a coarse garment proxy $V^c$. Inspired by~\cite{wang2019learning,holden2019subspace}, we adopt a \textit{Joint2Coarse} network, which generates the coarse garment geometry from the motion of the 3D joints as shown in Fig.~\ref{fig:coarse}. We represent the body pose at each frame as the global positions of $J$ selected body joints (see Fig.~\ref{fig:coarse}). Since the dynamics of the garment depends on the body motion in the past several frames, we combine the pose at the current frame and the past $K$ frames as $\hat{M}_t = M_{(t-K):t} \in R^{(K+1)\times J\times 3}$ to define a motion descriptor for frame $t$. Note that in $\hat{M}_t$, all the joint positions are represented relative to the root position at frame $t$. 

\begin{figure*}[t!]
\begin{overpic}[width=\textwidth]{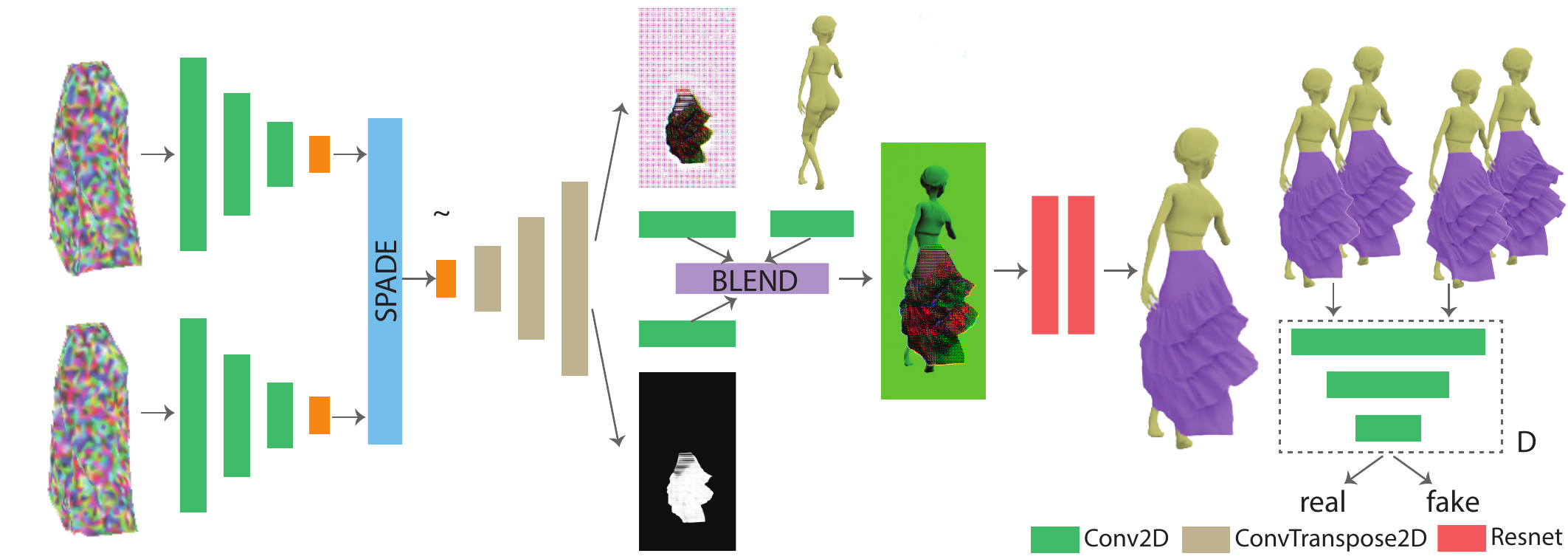}
    \put(2,33){$Q^p_{t-1}$}
    \put(2,15){$Q^p_t$}
    \put(20,28){$Z^p_{t-1}$}
    \put(20,12){$Z^p_t$}
    \put(20,28){$Z^p_{t-1}$}
    \put(28,20){$\tilde{Z}^p_t$}
    \put(38,1){$A^p_t$}
    \put(38,33){$U^p_t$}
    \put(54,33){$B^p_t$}
    \put(76,29){$R^p_t$}
    \put(82,34){$(R^p_t,I^p_{t-1})$}
    \put(92,34){$(I^p_t,I^p_{t-1})$}
    \put(52,8){$(1-A^p_t)B^p_t + A^p_t U^p_t)$}
    \end{overpic}
  \caption{Given the neural descriptor maps $Q^p_t$ and $Q^p_{t-1}$ of two consecutive frames, our neural rendering network first maps them to latent codes $Z^p_t$ and $Z^p_{t-1}$. $Z^p_t$ is further normalized to $\tilde{Z}^p_t$by a SPADE layer which utilizes the temporal dependency between the latent codes of the frames. We then generate both the appearance of the garment $U^p_t$ and a mask $A^p_t$ which we use to blend $U^p_t$ with the background image $B^p_t$. We further refine the composition and generate the final rendering $R^p_t$. The discriminator $D$ classifies a pair of renderings as `real' or `fake'.}
  \label{fig:NetModel}
\end{figure*}
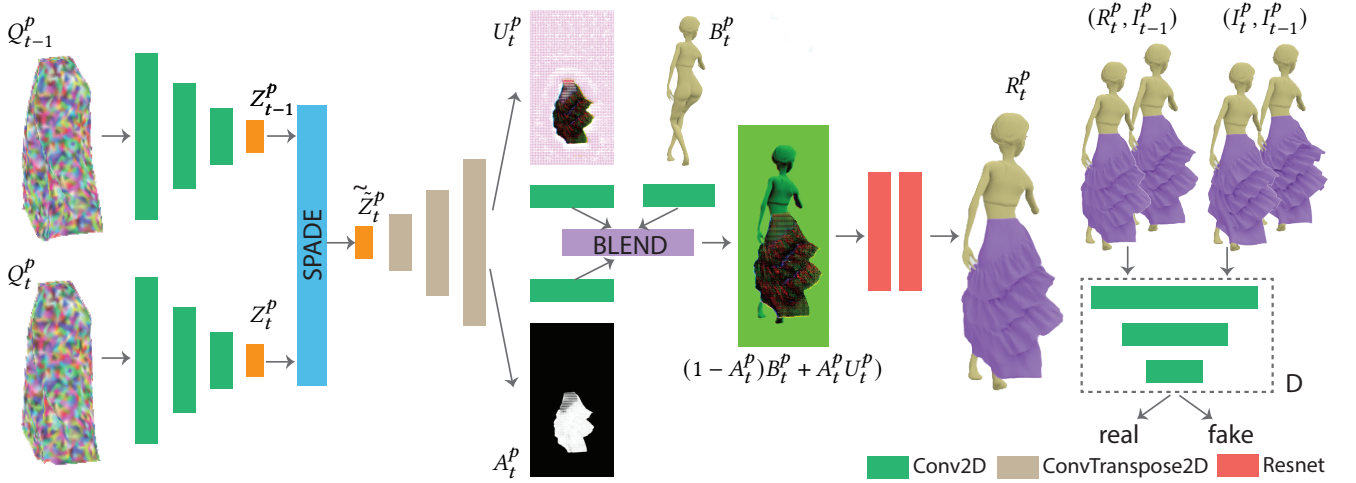

We first adopt an auto-encoder (cf., \cite{wang2019learning})  $\upzeta_D(\upzeta_E(\cdot))$  to learn a compact representation of the coarse template $V^c$ as it deforms under varying body motions by optimizing for $\upzeta_D(\upzeta_E(V^c)) \approx V^c$. We also learn a motion encoder network, $\upzeta_M (\cdot)$, that maps a given motion descriptor $\hat{M}_t$ to the corresponding latent representation of the coarse template at frame t, i.e., $\upzeta_M(\hat{M}_t) = \upzeta_E(V^c_t)$. At test time, given a new motion descriptor, we can generate the coarse garment geometry by applying $V^c_t = \upzeta_D(\upzeta_M(\hat{M}_t))$. While this approach synthesizes plausible dynamics of the coarse garment, it does not explicitly handle potential penetrations between the garment and the body. We find that utilizing additional post-processing steps to resolve such collisions as in previous work~\cite{wang2019learning,guan2012drape} is not necessary in our setting. The generated coarse template is used as a guide to drive the dynamic neural rendering which is prone to such collisions as shown in Fig.\ref{fig:NetModel}.

%\textbf{\emph{Template motion.}} The coarse template motions are generated by the motion invariant encoding network (MIEN). As described in~\cite{wang2019learning}, we assume that the dynamic garment is significantly related to the current and  the past $K$ frames of character body motions. We select $J$ body joints to describe frame $t$ motion with the 3D joint positions $M_t\rightarrow \{m_t^j\}$. The coarse template garment is a low resolution mesh $V_t$  that stores the 3D position of the vertices at frame $t$. As shown in Fig.~\ref{fig:MIEN}, with a pair of the motion $M$ and the template mesh $V$ as input, the MIEN first uses MLP encoder to separately encode $M$ and $V$ to the compact latent code $M_D$ and $V_D$ . The motion invariant latent vector $z$ is computed by associated blending the motion specified weights in multiple sub-networks $F_E(V_D,M_D)$. Given the latent vector $z$ and a motion descriptor $M$, the decoder outputs a plausible clothing mesh $V$ via $N_D(F_D(z|M))$. More details please refer to Fig.~\ref{fig:MIEN} and \cite{wang2019learning}. The MIEN is trained with synthetic data from a physically-based simulator, which inevitably introduces ‘noise’ to the dataset. This leads no guarantee on a collision-free garment prediction. In \cite{wang2019learning}, they use a post processing refinement to remove the intersections between the garment and the character body. However, in our system, the coarse template is only treated as a shape guidance for the target rendering, so we don’t need to solve the collision problem in this stage.

\subsection{Dynamic Neural Garment}\label{subsec:DNG}
Given the generated coarse garment sequence, $V^c_t$, we present a dynamic neural rendering approach to synthesize the final target garment appearance from a desired viewpoint. We explore the fact that the coarse template provides spatial-temporal correspondences across the motion sequence and learn neural features directly on the coarse garment to encode the style-specific appearance changes between the coarse and the target garment. The global neural features learned on the coarse garment are dynamically displaced across frames as the garment deforms. Importantly, to encode pose-specific dynamic appearance changes, we also condition the neural renderer on motion features extracted from the underlying body motion. We next describe the details of the neural and motion features, the architecture of the neural rendering network, and the loss function used to train it.

\textbf{\emph{Learnable neural features.}} In order to encode the style specific appearance relationship between the coarse and target garments, we learn a $d$-dimensional neural texture $F$ for the coarse garment $V^c$. In order to avoid overfitting to a specific texture resolution, our neural texture is represented in a multi-scale manner~\cite{thies2019deferred}. In our implementation, we learn a neural texture hierarchy of $4$ layers in our experiments. Note that, unlike \cite{thies2019deferred}, we do not encourage the first three channels of the neural texture to be similar to the RGB color of the corresponding pixel in the target image since our coarse template differs from the target garment both from a geometric and appearance perspective. Given the deformed coarse garment $V^c_t$ at each frame with known uv-coordinates, we sample the neural texture from the given viewpoint $p$ to generate a neural feature image $F^p_t$.

\textbf{\emph{Motion features.}} While the neural features learned over the coarse garment are geometrically displaced as the garment deforms, they are global and do not encode pose-specific appearance changes of the target garment. Hence, we condition our neural renderer on additionally defined motion features. Since our neural renderer is not restricted to a specific view, our motion features should be invariant to different camera parameters. To achieve this, for each pixel in the target rendering at frame $t$ from a specific viewpoint $p$, we first compute the 3D position of the corresponding location on the surface of the deformed coarse garment via barycentric coordinate, which we call $v^i_t$. We form a $J$ dimensional pose feature image $\hat{S}_t^p$ where for each pixel $i$ the $j^{th}$ channel encodes the distance of the corresponding vertex to the $j^{th}$ joint:
\[
    \hat{S}_t^p(i, j):= exp\left(-{\|v_t^i-M_{t}^j\|^2}/{\sigma}\right).
\]
In order to incorporate the motion information in the past $L$ frames, we concatenate the pose feature images, $\hat{S}_{(t-L):t}^p$, along the channel dimension to construct a motion feature image $S_t^p$ with $J \times (L+1)$ channels.

\textbf{\emph{Neural descriptor map.}} We concatenate the motion feature image $S^p_t$ and the neural feature image $F^p_t$ to form a final neural descriptor map $Q^p_t:=[F^p_t \;\; S^p_t]$ for each frame $t$. Our renderer network takes as input the features $Q^p_t$ along with the corresponding background image $B^p_t$ and synthesizes the final rendering of the target garment, as we will describe next.

\subsection{Rendering Network} \label{subsec:RenderingNet}
Our rendering network $G$ is mainly composed of a temporally coherent encoder and a layered based decoder. Given the neural descriptor maps of two consecutive frames $Q^p_t$ and $Q^p_{t-1}$, we first encode them into the latent space resulting in latent codes $Z^p_t$ and $Z^p_{t-1}$, respectively. We utilize a spatially adaptive normalization layer (SPADE)~\cite{park2019semantic} to normalize $Z^p_t$ into $\tilde{Z}^p_t$ conditioned by $Z^p_{t-1}$. This helps to implicitly achieve temporal consistency between the latent codes of consecutive frames. The layered based decoder takes as input $\tilde{Z}^p_t$ along with the background image $B^p_t$ and synthesizes the appearance of the target garment on the character. In order to ensure further temporal consistency between consecutive frames, we introduce a patch-based temporal-and-spatial discriminator to train our rendering network. We show the architecture of our network in Fig.~\ref{fig:NetModel}.

\textbf{\emph{Temporally coherent encoder}.} 
As shown in Fig.~\ref{fig:NetModel}, the encoder component of our network consists of a set of convolutional layers that encode the input neural descriptor maps $Q^p_t$ into latent codes $Z_t^p$.  We define a SPADE block which is used to normalize the signal conditioned with the concatenation of $Z_{t}^p$ and $Z_{t-1}^p$. This normalization with respect to the previous frame enables to extract the temporal dependency between the consecutive frames at every spatial location. Our SPADE block first normalizes the input feature map using scale $\gamma$ and bias $\beta$ parameters learned from $[Z_{t}^p|Z_{t-1}^p]$ with respect to both each channel $c$ and spatial location $(x,y)$ as described in~\cite{park2019semantic}. Specifically, for an input signal $\mathbf{w}$, we obtain the normalized $\tilde{\mathbf{w}}$ as:
\begin{equation*}
    \tilde{\mathbf{w}}[c,x, y]:=\mathbf{SPADE}(\mathbf{w})=\gamma_{c,x,y}\frac{\mathbf{w}[c,x,y]-\mu_c}{\sigma_c}+\beta_{c,x,y},
\end{equation*}
where $\mu_c$ and $\sigma_c$ are respectively the mean and standard deviation of $\mathbf{w}$ in channel $c$. We apply our SPADE blocks to $Z_{t}^p$ in a residual style. Specifically we obtain the SPADE-normalized latent code $\tilde{Z}_t^p$ as:
\begin{equation*}
    \tilde{Z}_t^p=\mathbf{SPADE}(\mathcal{I}(\mathbf{SPADE}(\mathcal{I}(Z_t^p))))+\mathbf{SPADE}(\mathcal{I}(Z_t^p)),
\end{equation*}
where $\mathcal{I}(\cdot)$ refers to an instance normalization layer.

%\meng{I think we should make $\gamma$ and $\beta$ more clear: $\gamma$ and $\beta$ are 2D convolution computed on the $[Z_t^p|Z_{t-1}^p]$ , represented as the pixel-and-channel-wise shifting and scaling and next used to normalise $Z_t^p$ . }

\textbf{\emph{Layered based decoder}.} The architecture of our decoder consists of a set of ConvTranspose2D layers symmetric to the encoder. Given $\tilde{Z}_t^p$, these layers first generate a feature image $U_t^p$ with the same width and height as the target rendering. The provided latent code and hence $U_t^p$ captures the appearance and dynamics of the target garment only. In order to composite the garment with the character body and the background, we introduce a layer blending module to blend $U_t^p$ with $\hat{B}_t^p$, the features extracted from the provided background image $B_t^p$. Specifically, we generate a mask $A_t^p$ and apply the learned mask to blend the foreground and the background by $(1-A_t^p) \cdot \hat{B}^p_t+A^p_t\cdot U_t^p$. We finally apply two residual convolution blocks to refine the result and generate the final rendering $R_t^p$.

\subsection{Loss Function}\label{subsec:LossFunction}
We train our dynamic neural rendering component to learn the parameters of the network $G$ and the neural texture $F$ jointly in an end-to-end manner. Given the final rendering $R_t^p$ synthesized by the network and the corresponding ground truth image $I_t^p$, we first consider an $L_1$ loss with respect to colours as well as multi-layer features of the pretrained network VGG:
\begin{equation*}
    L_{percept} :=\sum_i\|VGG^i[R^P_t]-VGG^i[I^P_t]\|_1 +\|R_t^P-I_t^P\|_1.
\end{equation*}

\textbf{\emph{Temporal-and-spatial discriminator}.} Furthermore, we adapt an adversarial loss to ensure the spatio-temporal plausibility of the renderings. Specifically, we train a discriminator network $D$ that takes consecutive frames as input and classifies them as real or fake. We provide two types of input to $D$. First, we provide $[R_t^P, I_{t-1}^P]$ as input while the samples $[I_t^P, I_{t-1}^P]$ provide real examples. We also provide $[I_{t+1}^P, R_{t}^P]$ as input where samples $[I_{t+1}^P, I_{t}^P]$ provide the real examples. These samples ensure that the generated rendering at frame $t$ is temporally consistent with the ground truth previous and next frames. We adopt a patch-based discriminator $D$ (cf., \cite{isola2017image}) to model high frequency details by restricting the attention to local patch structures. To further ensure temporal consistency, given two renderings $[I^P_t,I^P_{t-1}]$ of the neighboring frames, $D$ tries to identify if each $N \times N$ patch is real or fake both in the spatial domain as well as the temporal domain along the channel direction when they are concatenated. %~\cite{isola2017image}. %\duygu{for the discriminator, do we sample N by N patches both on the original rendering and also on concatenated renderings of two consecutive frames along the channel dimension?} \meng{In PatchGAN, the $N \times N$ is implicitly organized by 2D convolution as mentioned in ~\cite{isola2017image}}

The overall loss function for $D$ to minimize is: 
\begin{align*}
    &L_{D_1}=-log(D[I_t^P, I_{t-1}^P])-log(1-D[R_t^P, I_{t-1}^P]), \\
    &L_{D_2}=-log(D[I_{t+1}^P, I_t^P])-log(1-D[I_{t+1}^P, R_t^P]), \\
    &D := \argmin_D[L_{D_1}+L_{D_2}].
\end{align*}
For the rendering network, the adversarial loss is defined as:
\begin{equation*}
    L_{GAN}=-log(D[R^P_t,I_{t-1}^P])-log(D[I_{t+1}^P, R_t^P]).
\end{equation*}

We also use feature matching in multiple discriminator layers $D^i[*]$ to enforce similarity across different scales of features:
\begin{align*}
   L_{feat} =&\sum_i\|D^i[R^P_t,I^P_{t-1}]-D^i[I^P_t, I^P_{t-1}]\|_1 \\
   &+ \sum_i\|D^i[I^P_{t+1}, R^P_t]-D^i[I^P_{t+1}, I^P_t]\|_1.
\end{align*}

Given the different losses, we learn the weights of network $G$ as well as the neural features $F$ that minimize:
\begin{equation*}
    G, F := \argmin_{G,F}[\lambda_1 L_{feat}+\lambda_2 L_{percept}+\lambda_3 L_{GAN}].
\end{equation*}
Following SPADE \cite{park2019semantic} and SEAN \cite{zhu2020sean}, we set $\lambda_1=5, \lambda_2=10, \lambda_3=0.5$ in our experiments.

\subsection{Post-processing}
While $G$ consists of a layering module to composite the synthesized garment with the target body, occasionally we observe that the depth ordering between the arms which are highly dynamic and garment types such as skirts is not correct. To resolve these, we adopt a simple heuristic through an image layer re-composition approach. We first identify the arm regions in the character body renderings and the garment region in the final renderings. If the arm is closer to the camera (i.e., close or in front of the hip) but is occluded by the garment in the final rendering, we bring it (i.e., layer it) to the front.

\section{Results and experiments}
\label{sec:experiments}
In this section, we evaluate our approach with various scenarios and show that our method outperforms existing character appearance synthesis approaches qualitatively and quantitatively. We show that our method generalizes across unseen motion, novel camera view, and adapts to novel body shape, background and environment illumination with an efficient fine-tuning strategy.

\subsection{Data generation}
To train our method, we establish a synthetic dataset by first creating virtual avatars using Adobe Fuse. We rig and animate each avatar via Mixamo~(\url{https://www.mixamo.com/}) to generate a training motion sequence of $850$ frames. Next, we run a physically-based simulation using Marvellous Designer~\cite{md} to generate the ground truth mesh sequence for both the coarse template and the target garments. Specifically, we use a particle distance of $30mm$ when simulating the coarse template and a distance of $10mm$ when simulating the target garments to capture high-frequency details. Since we know the ground truth body joint positions, we can also compute the motion descriptor used by both the coarse garment synthesis (Section \ref{subsec:CoarseTemplate}) as well as the motion features used by the dynamic neural rendering (Section \ref{subsec:DNG}) of our method. The coarse templates are used to train our \emph{Joint2Coarse} network to provide the neural feature proxies. To ensure generalisation of the rendering network across camera views, at each frame, we randomly position 10 cameras on a circle around the avatar. Under each view $P$, we generate the required ground truth input and output maps for 3 consecutive frames $\{t-1, t, t+1\}$ as required in the temporal adversarial loss function (Section~\ref{subsec:LossFunction}). We set up the rendering scene in Blender with an HDRi environment map. In order to keep the illumination consistent across different views, we rotate the HDRi map to keep a fixed relative position with respect to the camera. In total, our training set for the rendering network includes $8500$ samples.
%\meng{I did some change in this paragraph. Please take a pass.}

Our dataset consists of three coarse garment templates: a long skirt ($2287$ vertices), a short skirt ($1756$ vertices), and a full-body dress ($5087$ vertices). We also define $5$ different target garments: (i)~a multi-lace skirt with $48389$ vertices; (ii)~a double-layer skirt with $33196$ vertices; (iii)~a tango skirt with $26718$ vertices; (iv)~a short hem skirt with $56657$ vertices; and (v)~a full-body combination composed of a delicate dress skirt, blouse, ropes, and strings with a total of $51941$ vertices. For the short hem skirt, we use a particle distance of $5mm$ when training our method to produce rich folds, while the others use $10mm$ spacing. For the first three types of skirts, we use the same coarse long skirt template. For each coarse template and target garment combination, we train a specific neural rendering network along with a new set of neural textures. 

\begin{figure}[!b]
  \includegraphics[width=\columnwidth]{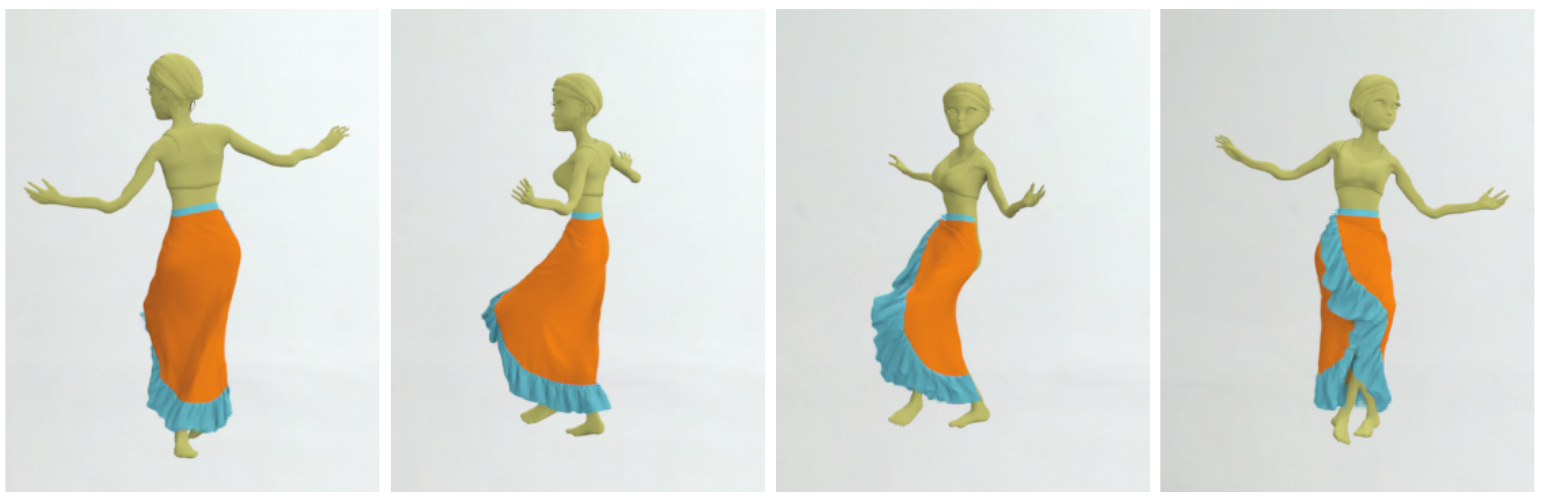}
  \caption{Our method can synthesize the target garment appearance from different viewpoints not seen during training.}
  \label{fig:diffViews}
\end{figure}

\begin{figure}[!t]
  \includegraphics[width=\columnwidth]{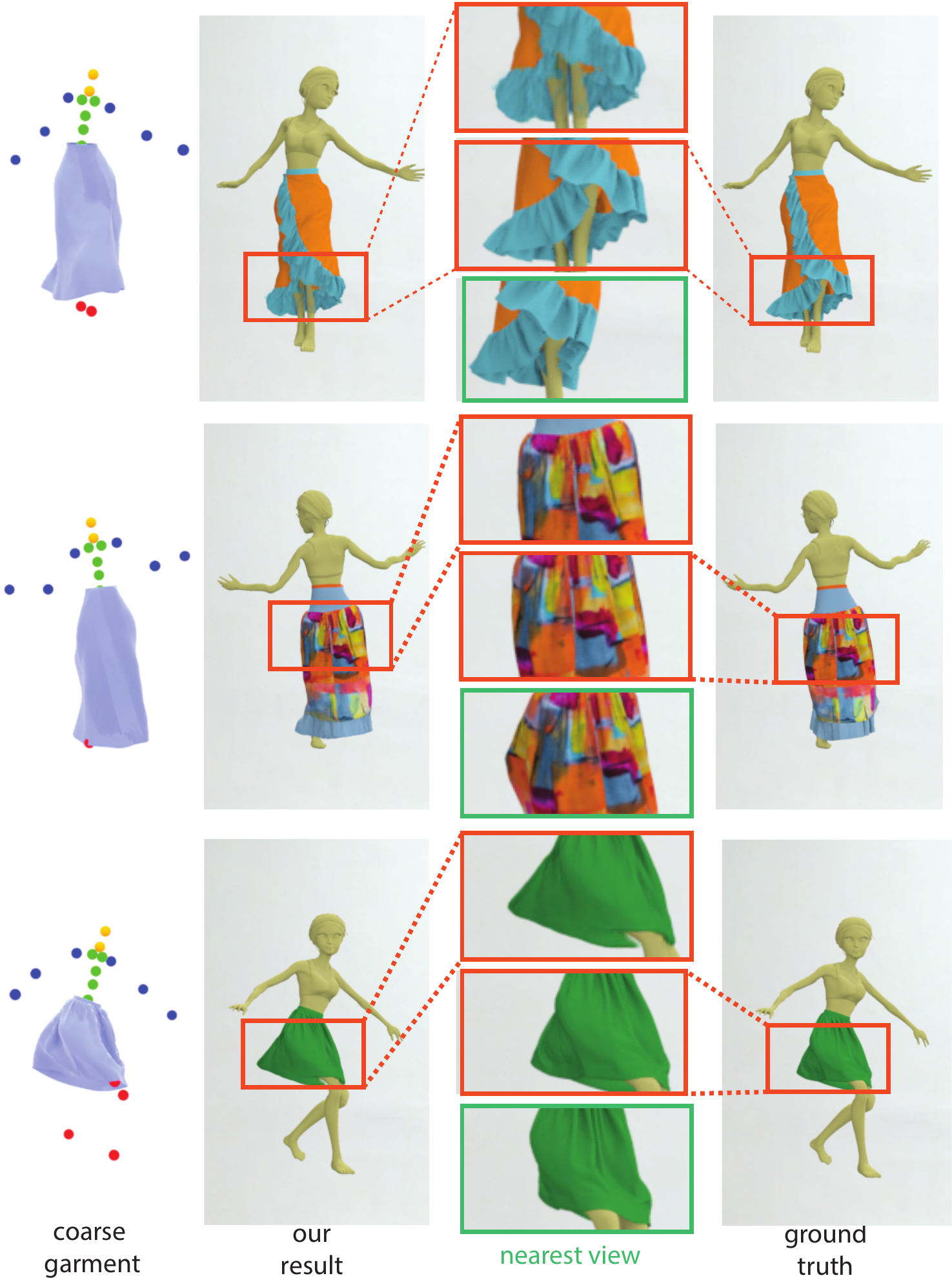}
  \caption{We test our method on the training motion sequence on unseen viewpoints. For each example, our result and the ground truth. Notice how our results are closer to the ground truth in the highlighted regions compared to the nearest views in the training data shown in green.}
  \label{fig:seenMotion}
\end{figure}

\subsection{Implementation details}
Our \emph{Joint2Coarse} network consists of a shape encoder $\upzeta_E(\cdot)$, a shape decoder $\upzeta_D(\cdot)$, and a motion encoder $\upgamma(\cdot)$. The shape encoder $\upzeta_E(\cdot)$ takes the $v \times 3$-dimensional vector as input ($v$ is the number of vertices in the template) and maps it into a latent space through 6 fully connected layers with output dimensions gradually decreasing to 2048, 1024, 512, 256, 128, and 64. The shape decoder reverses this process symmetrically. We build a motion descriptor consisting of $J = 19$ body joint positions of the current as well as the past $34$ frames (a total of $K=35$ frames ), resulting in $\hat{M}_t \in R^{35\times 19\times 3}$. We flatten the motion descriptor and obtain a 969-dimensional vector as the input of $\upgamma(\cdot)$. The motion descriptors are then mapped into the 64-dimensional latent space through 4 fully connected layers with output dimensions gradually decreasing to 512, 256, 128, 64. We adopt ReLU activations for all the layers and dropout connections with a rate of $0.05$. The encoders and the decoder are trained by the RMSprop algorithm with a learning rate of $10^{-3}$. 
%The training takes about 10 mins to converge for our experiments.

As shown in Fig. \ref{fig:NetModel}, the input of the rendering network is dynamic pairs of neural descriptor maps $Q_t^P$ and $Q_{t-1}^P$. Each map consists of per-pixel neural features of dimension $64$ sampled from a neural texture hierarchy composed of 4 levels, each with 16 channels. In addition, the maps also include per-pixel motion features with dimension $ J \times L$, where $J=19$ (the 19 body joints), and $L=5$ (the $5$ sampled previous frames with an interval of 2 to cover a period of 10 frames. The encoder first down-samples the resolution of the input maps from $512 \times 512$ to $16 \times 16$, and increases the feature dimension to 512 by 2D convolution layers each with a Leaky ReLU activation. In the SPADE module, we use a SPADE residual block \cite{park2019semantic}, keeping the feature size to be $16 \times 16 \times 512$. Then we use 2D convTranspose layers to increase the output resolution to $512 \times 512$ but still maintain high dimensional features of size 32. After blending the background image features and the garment rendering features, we apply $2$ residual convolution blocks to output the final rendering images with a size of $512 \times 512$ in RGB color space. We adopt a patch size of $70 \times 70$ \cite{isola2017image} in our temporal-and-spatial discriminator. In the training stage, we use the Adam optimiser, with the default parameters $\beta_1=0.9, \beta_2=0.999$, and a learning rate of $lr_1=1.e-4$ for the rendering network including the neural texture optimization, and $lr_2=3.e-4$ for the temporal-and-spatial discriminator. It took about 35 epochs to converge when training the rendering network with a batch size of 2.
%\tuanfeng{time and machine specifications}

\subsection{Results and evaluation}\label{subsec:results}
%In the experiments as shown in Figure x, we target at 3 well-designed skirts with a high resolution simulation of $10mm$ particle distance running in Marvelous Designer, that the number of mesh vertices are respectively $48,389$(multi-laces), $33,196$(double-levels), $26,718$(tango-style). The amount of mesh vertices dynamically varies along with the avatar body shape to ensure the detail wrinkles consistent with the body structure, for example, once the body shape become wider, the tango-style increases the number of vertices to $30,088$, as shown in Figure x. All those target skirts share the same coarse skirt as the template, that is generated with a particle distance of 30mm, resulting 2,287 vertices. In the dirndl experiment, we use a simple dress with $2,327$ vertices as the coarse template, aiming at a complicated designed dirndl with $51,941$ composed of delicate dress skirt, blouse, roops and strings. 

We now evaluate the generalization ability of our method. While showing illustrative figures, we also provide a supplementary video to better evaluate the visual quality of the generated garments under motion.

\textbf{(i)~Unseen views.} For the motion sequence seen during training, we test how well our method works with different, unseen viewpoints as shown in Fig.~\ref{fig:diffViews} and~\ref{fig:seenMotion}. In Fig.~\ref{fig:seenMotion}, for each test view, we show the training samples with the nearest viewpoint. As can be seen in the supplemental video, compared to the nearest training samples, our method generates temporally coherent sequences for the unseen views.

\begin{figure}[!t]
  \includegraphics[width=0.9\columnwidth]{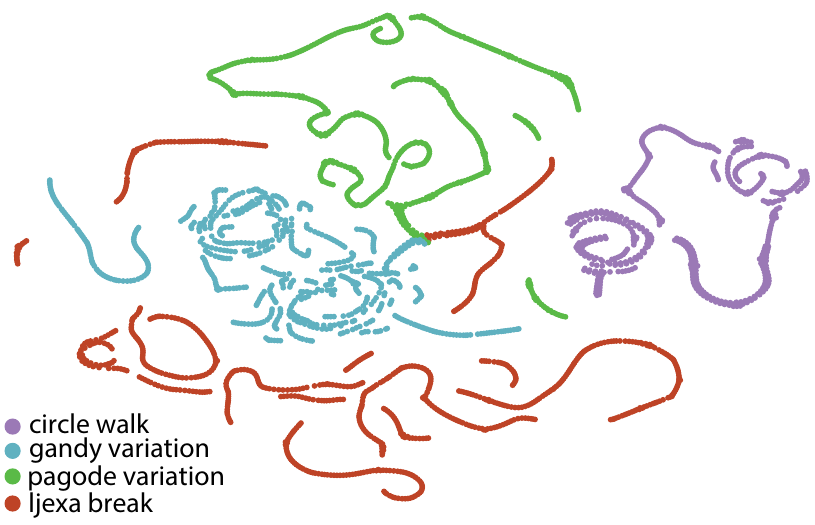}
  \caption{We visualize the distribution of motion sequence used as training data (\emph{ljexa break}) and testing data (\emph{pagode variation}, \emph{gandy variation}, and \emph{circle walk}) via t-SNE~\cite{maaten2008visualizing}.}
  \label{fig:tsne}
\end{figure}
\textbf{(ii)~Unseen motions.} Next, we evaluate the generalization ability of our method across unseen motion sequences. In Fig.~\ref{fig:tsne}, we first visualize the distribution of the body poses observed both in training and testing motion sequences via t-SNE~\cite{maaten2008visualizing} over the 3D body joint positions with respect to the root joint (i.e., in local coordinates). We use the \emph{ljexa break} dancing motion sequence to train our network and test on the \emph{pagode variation}, \emph{gandy variation}, and \emph{circle walk} sequences. Our method generalizes well to motion sequences covered by the distribution of the poses seen during training. As with many deep learning approaches, as the distribution of testing motions (e.g, circle walk) becomes significantly different than the training data, we observe a drop in performance (e.g., flickering across frames).

% Additionally, in Fig.~\ref{fig:real},  we test our method on the motion sequences extracted from real capture. We use xxx (method) to extract motion joints from a video as input of \emph{Joint2Coarse} network to compute the coarse templates. With the neural features attached to the coarse templates, our rendering network produces the rendering frames of the virtual avatar dressing our target garment driven by the video motion.
%\begin{figure*}[!h]
%  \includegraphics[width=\textwidth]{fig/unseen2.pdf}
%  \caption{We test our method on unseen motion sequences.}
%  \label{fig:unseenMotion2}
%\end{figure*}

\begin{figure}[!h]
  \includegraphics[width=\columnwidth]{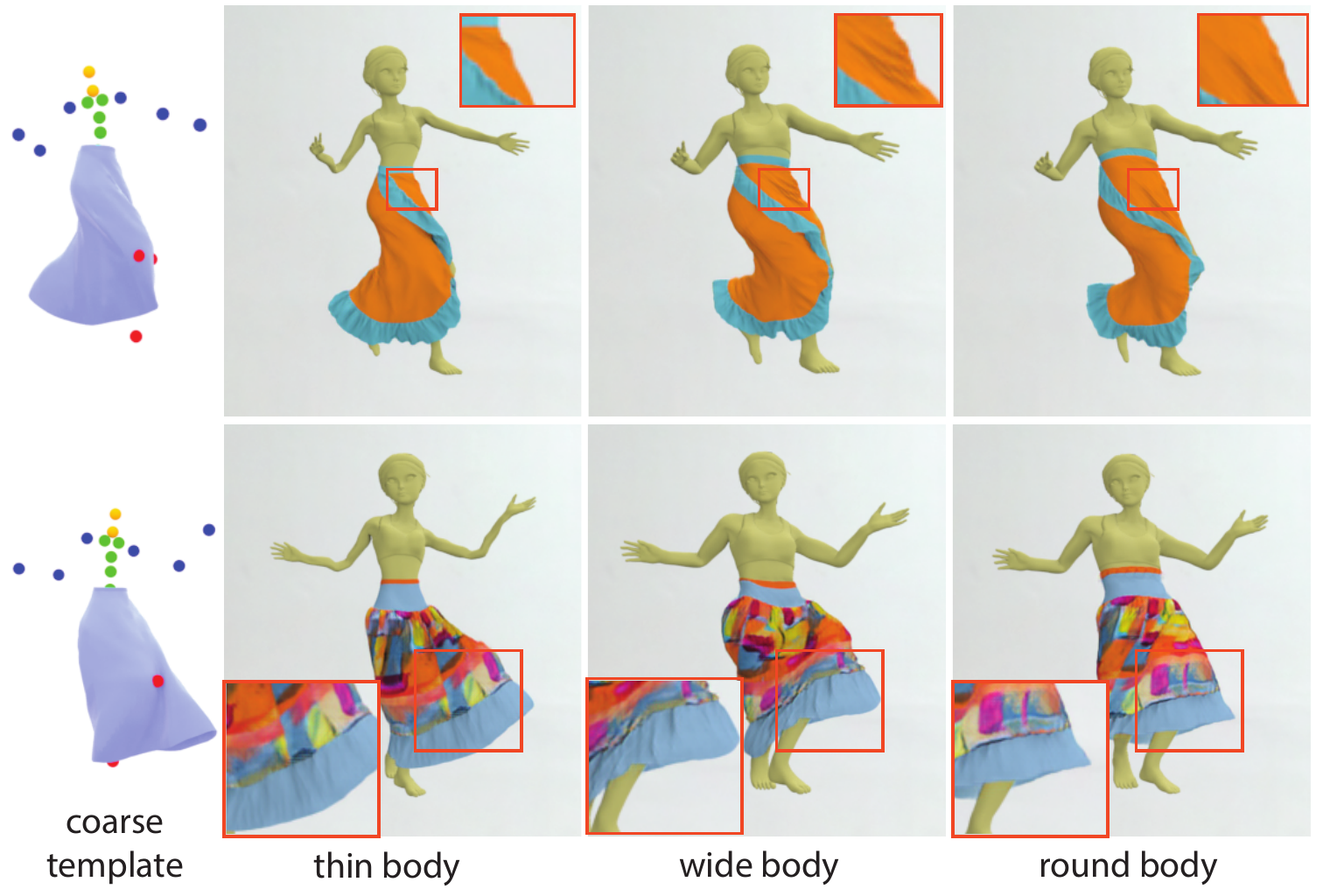}
  \caption{Given a coarse garment motion, we synthesize how the target garment would fit different body shapes.}
  \label{fig:diffBody}
\end{figure}

\begin{figure*}[!t] 
  \includegraphics[width=0.95\textwidth]{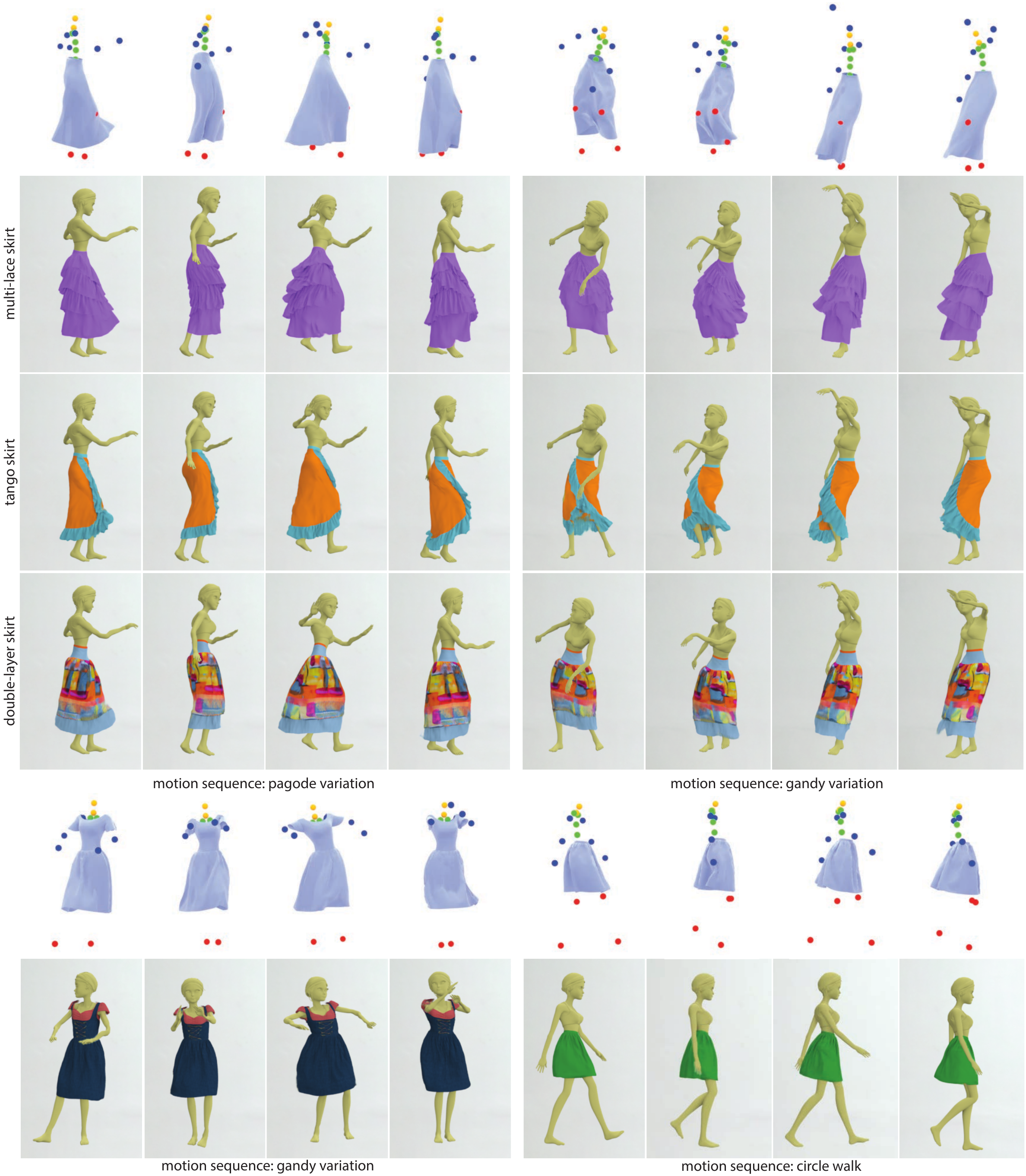}
  \caption{We evaluate our method on different garment types with unseen motion sequences. We note that we learn a garment specific neural texture and renderer.}
  \label{fig:unseenMotion}
\end{figure*}

\textbf{(iii)~Different body shapes.} We also evaluate our method with varying body shapes, as shown in Fig.~\ref{fig:diffBody}. Given neural textures and the renderer learned for a particular body shape, e.g., a thin body, we fine-tune the network for a new body shape. We use the same coarse template of the thin body. We find that fine-tuning the network reduces the number of training iterations to $20-30\%$ compared to training from scratch.

\begin{figure*}[!h]
  \includegraphics[width=\textwidth]{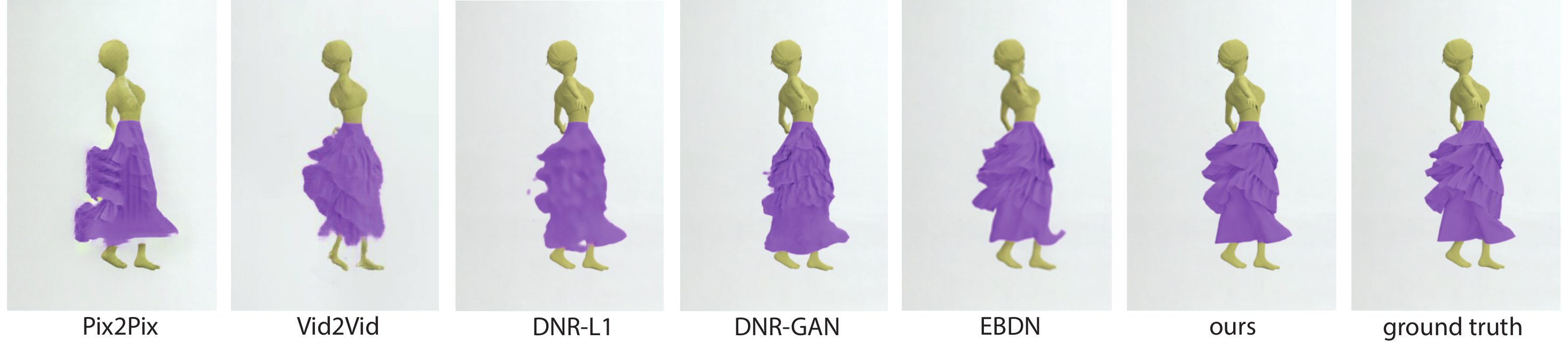}
  \caption{We compare our method to image and video translation methods, Pix2Pix~\cite{isola2017image} and Vid2Vid~\cite{wang2018vid2vid} as well as DNR~\cite{thies2019deferred}, a deferred neural rendering approach, and EBDN~\cite{chan2019everybody} that focuses on pose editing of humans. Our approach outperforms these baseline methods.}
  \label{fig:comparison}
\end{figure*}

\begin{figure}[!h]
  \includegraphics[width=\columnwidth]{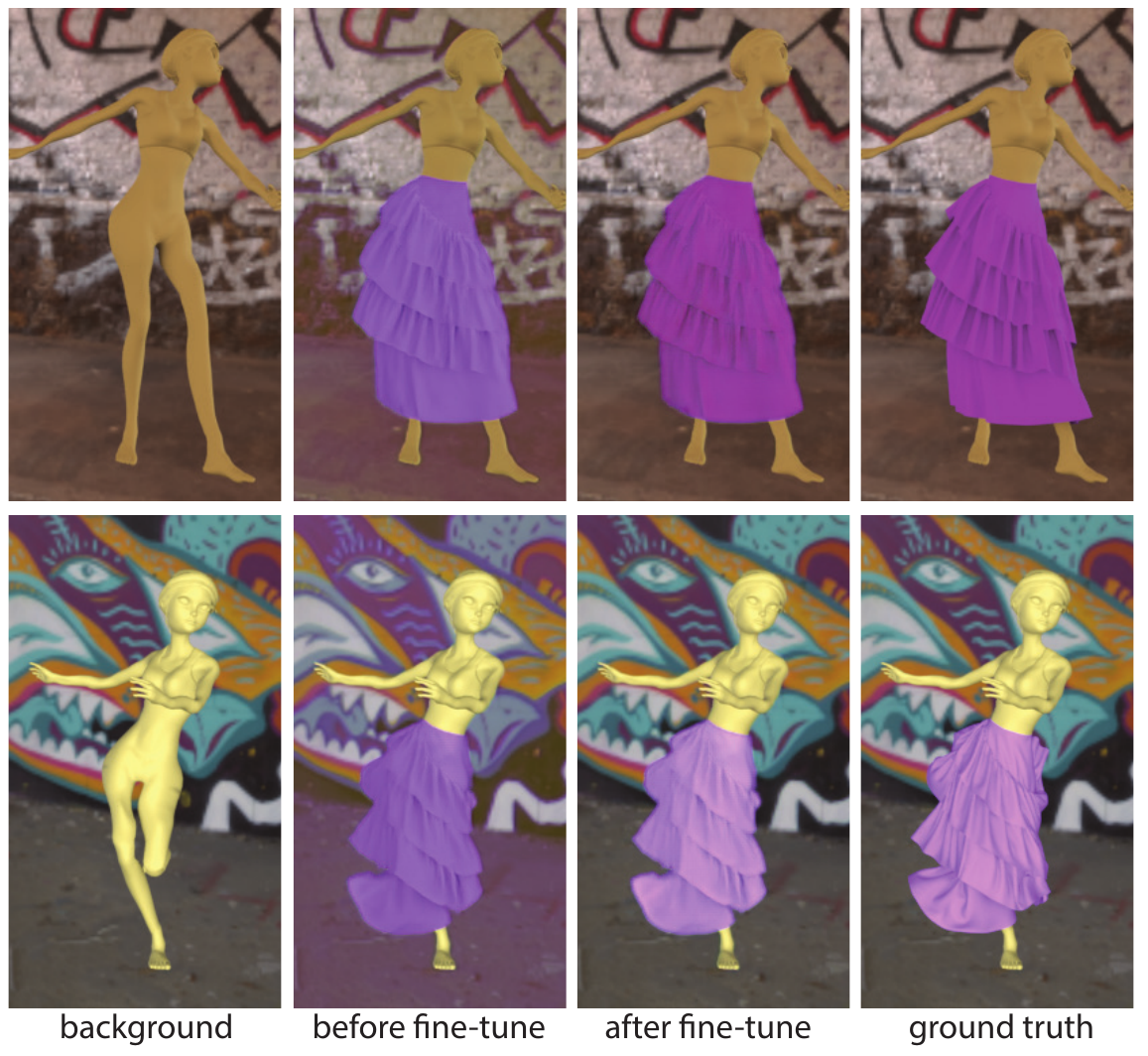}
  \caption{Our dynamic neural renderer can easily be fine-tuned with a few examples to generalize to different backgrounds showing various illumination conditions.}
  \label{fig:background}
\end{figure}

\textbf{(iv)~Different background.} Finally, we extend our method to synthesize images with various background images demonstrating different illumination conditions by fine-tuning the last two layers of our layered based decoder jointly with the discriminator as shown in Fig.\ref{fig:background}. Our fine-tuning strategy allows us to achieve plausible results by training with two example images rendered with novel background and environment illumination, which takes 100 training iterations to converge with a batch size of 2.
%which significantly reduces the training time to xx seconds with yy epochs. (\meng{I think we don't need to mention the training time because of my un-optimized code. For background fine-turn, we take 100 iterations running on the 2 image (batch size=2) and stop the training before over fitting on the 2 example images. })

\begin{figure}[!h]
  \includegraphics[width=\columnwidth]{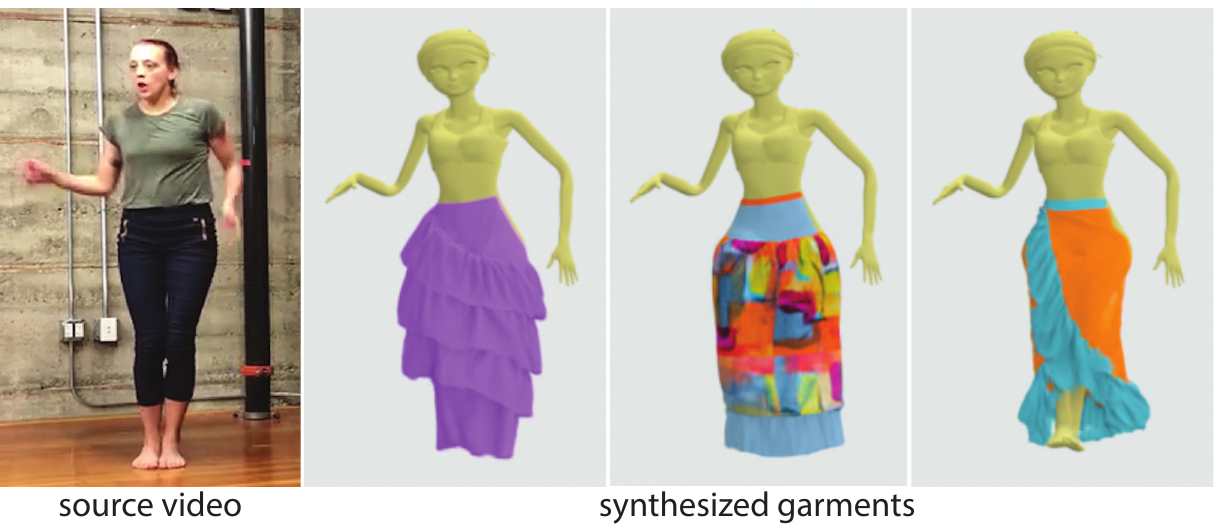}
  \caption{Given a source input video, we use the estimated motion of the actor to drive a 3D character while synthesizing plausible garment appearances.}
  \label{fig:real}
\end{figure}

\textbf{Real capture.} Our method can be combined with 3D body motion estimation methods to drive a 3D character wearing a desired garment and synthesize the corresponding dynamic appearance. In Fig.~\ref{fig:real}, we provide an example where we have used the method of Rempe et al.~\cite{RempeContactDynamics2020} to extract the 3D human body motion from the source video.

\textbf{Computational performance.} Our method learns to simulate and render a garment efficiently given a body motion sequence. Once trained, our method takes $40$ ms per frame to generate the coarse template mesh and 20ms seconds per frame to render the final image. We run our experiments on a PC with Intel Xeon CPU E5-1650, 64GB of memory, and an NVIDIA GeForce GTX 1080Ti graphics card. In comparison, in a professional garment simulation tool such as Marvellous Designer, each frame of the simulations takes $1.58$ seconds for the multi-lace skirt, $1.05$ seconds for the tango skirt, $1.10$ seconds for the double-layer skirt, $2.03$ seconds for the short hem skirt, and $2.31$ seconds for the full-body dress. 
%\tuanfeng{rendering time?}

\subsection{Baseline comparisons}\label{subsec:baselineComp}
We compare our method against image and video translation methods, (a)~Pix2Pix~\cite{isola2017image} and (b)~Vid2Vid~\cite{wang2018video}, (c)~a human body reposing method Everybody Dance Now~\cite{chan2019everybody} (EBDN), and (d)~a closely related neural rendering method, Deferred Neural Rendering (DNR)~\cite{thies2019deferred}. 

For Pix2Pix~\cite{isola2017image}, we experimented with two variants: (i)~learning the mapping between the joints and the final appearance; (ii)~learning the mapping between the rendering of the coarse template and the final appearance. The first version provided unsatisfactory results. Hence, we only present the second version in our comparisons. 
For Vid2Vid~\cite{wang2018video}, we adopt Openpose~\cite{cao2019openpose} to extract 2D skeletons and Densepose~\cite{guler2018densepose} to extract dense UV maps for the character body as input. The extracted 2D skeleton is also used for EBDN to drive the motion synthesis. 
For DNR~\cite{thies2019deferred}, we use our generated coarse template as the underlying geometry for each frame and jointly learn a corresponding neural texture and a neural renderer as described in the original paper with the L1 loss. We also train the same network with adversarial loss as in our method (Section~\ref{subsec:LossFunction}).  All the methods are trained with the same training set as ours and tested with the same motion under different viewpoints as shown in Fig.~\ref{fig:comparison}. In Table~\ref{table:baseline-quant}, we provide a quantitative comparison and show that our method outperforms the baselines with seen motion. With the strongest baseline, EBDN, we also show a comparison with unseen motion in the supplementary video. Our method performs significantly better than the baseline in this case.

\begin{table}[h]
\caption{For quantitative comparisons, we show the mean ($\mu_\text{mse}$) and standard deviation ($\sigma_\text{mse}$) of the per-frame mean-square-error between different methods and the ground truth image; the Fréchet Inception Distance (FID)~\cite{heusel2017gans}; and the video based Fréchet Inception Distance (V-FID)~\cite{wang2018vid2vid} for the whole sequence generated by each method. The scale of the pixel value is $[0,1]$ for each channel.}
\label{table:baseline-quant}\small
\begin{tabular}{c||c|c|c|c|c|c} 
& Pix2Pix & Vid2Vid & DNR(L1) & DNR(GAN) & EBDN & Ours \\
\hline \hline
$\mu_\text{mse}$ & 0.591 & 0.734 & 0.204 & 0.254 & 0.204 & \textbf{0.116} \\
\hline
$\sigma_\text{mse}$ & 0.120 & 0.188 & 0.081 & 0.082 & 0.064 & \textbf{0.045} \\
\hline
FID & 75.77 & 96.21 & 71.12 & 36.41 & 38.31 & \textbf{11.39} \\
\hline
V-FID & 2.18 & 3.31 & 2.08 & 1.68 & 0.74 & \textbf{0.40} \\
\end{tabular}
\end{table}

\subsection{Ablation Study}

\begin{figure}[!h]
  \includegraphics[width=\columnwidth]{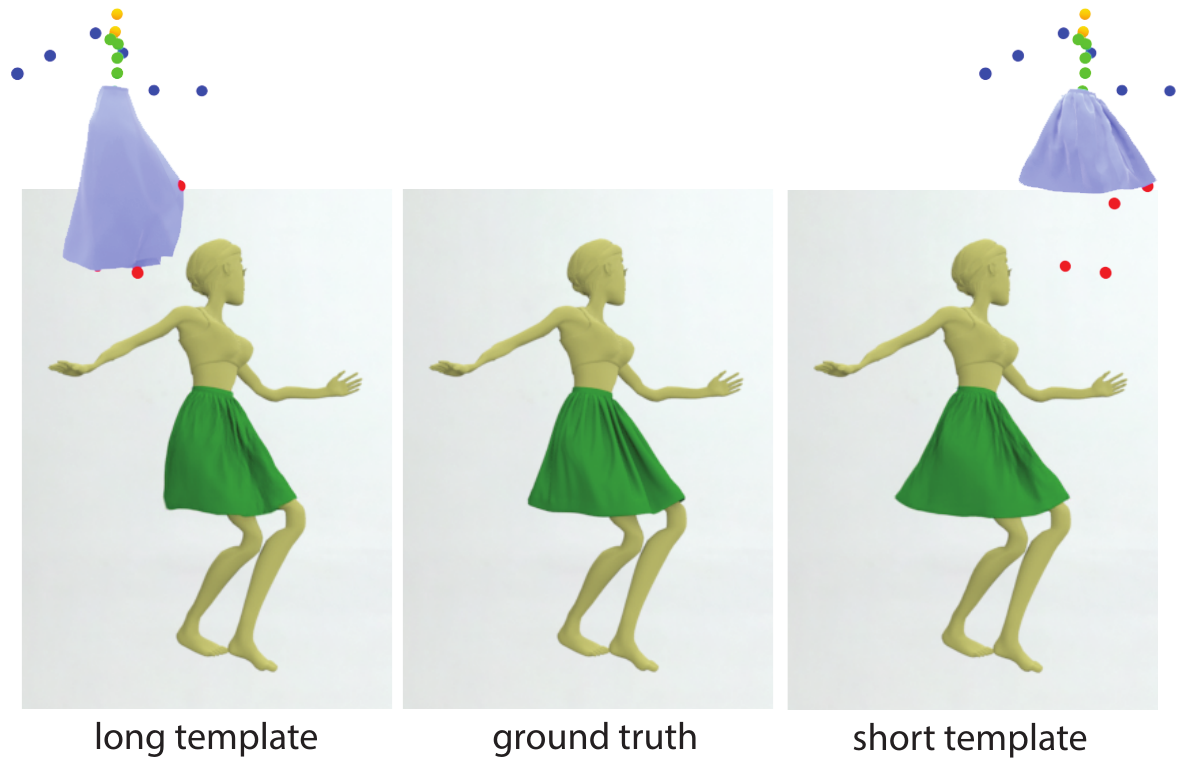}
  \caption{For a target short pleated skirt, we show the results of using a long and a short skirt template.}
  \label{fig:effectTemplate}
\end{figure}

\textbf{(i) Effect of the coarse template.} When choosing the coarse templates for a particular garment, we tend to choose a garment of a neutral type without geometric details such as laces or layering. While a coarse template with more geometric details similar to a specific target garment can potentially yield better results, it is hard to generalize across different target garments with the same template. Hence, we aim to achieve a balance. As seen in Fig.~\ref{fig:unseenMotion}, we use the same plain long skirt garment for three different skirt types, including a multi-lace skirt, a double-layer skirt, and a tango style skirt. For another short hem skirt, we experimented with using both the aforementioned long skirt template and a short skirt template. While the use of the short skirt template improves the quality of the results, the long skirt template still yields reasonable results as shown in Fig.~\ref{fig:effectTemplate} and Table~\ref{table:ablation-quant}.

\begin{figure}[!h]
  \includegraphics[width=\columnwidth]{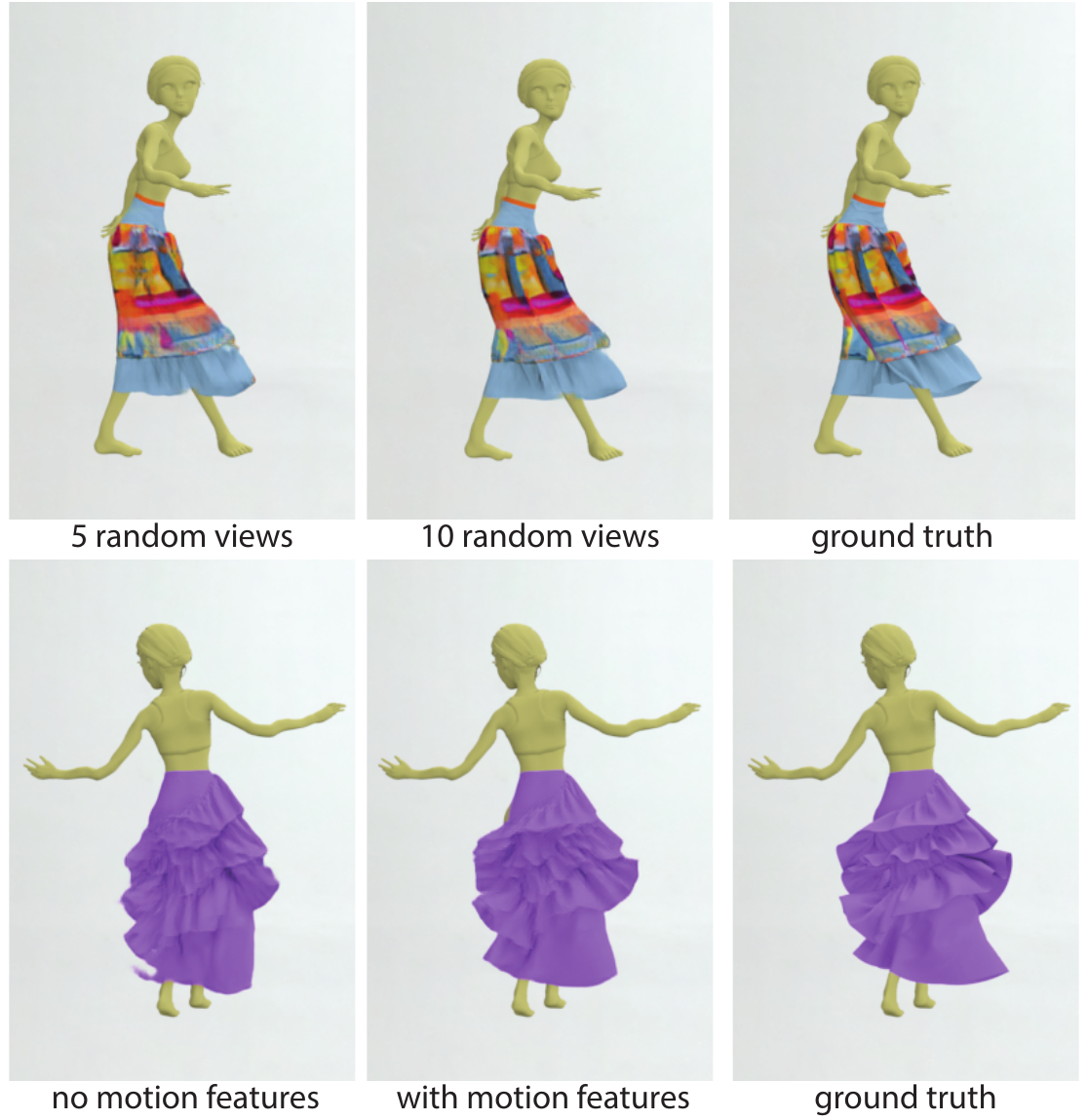}
  \caption{We perform an ablation study to demonstrate the importance of the number of views used during training as well as the use motion features representing the underlying body motion.}
  \label{fig:ablation}
\end{figure}

%>> neural descriptor input with / without motion features
\textbf{(ii) Effect of motion features.} To demonstrate the importance of using motion features ($S_t^P$) for dynamic neural rendering, we train the renderer with and without motion features. Our renderer is able to model the rough motion of the garment even without motion features due to the adoption of the coarse template. However, the results improve significantly and high-frequency details are better preserved when motion features are added as shown in Fig.~\ref{fig:ablation} and Table~\ref{table:ablation-quant}.

%>> train with 5 views per frame vs 10 views
\textbf{(iii) Effect of number of views used for training.} When training our dynamic neural rendering network, we sample 10 random views for each frame. Training with different camera views helps with the generalization across unseen views and unseen poses due to the relative position of the 3D character changes with respect to the sampled cameras. To evaluate the importance of the number of views used during training, we train the network with $5$ and $10$ different views for each training sample. As shown in Fig.~\ref{fig:ablation} and Table~\ref{table:ablation-quant}, increasing the number of views improves the visual quality of the results.

\begin{table}[h]
\caption{We show the Fréchet Inception Distance (FID)~\cite{heusel2017gans} for the experiments in our ablation study. We show that our approach achieves better performance compare to other alternatives.}
\label{table:ablation-quant}%\tiny
\scalebox{0.8}{
\begin{tabular}{c||c|c} 
double-layer skirt in Fig~\ref{fig:ablation} & 5 random views & 10 random views (ours)\\
\hline
FID score & 24.00 & 19.18 \\
\hline \hline
multi-lace skirt in Fig~\ref{fig:ablation} & no motion features & with motion features (ours)\\
\hline
FID score & 13.46 & 11.34 \\
\hline \hline
green skirt in Fig~\ref{fig:effectTemplate} & long template & short template\\
\hline
FID score & 8.34 & 7.57 \\

\end{tabular}
}
\end{table}

%\subsection{Application (Potential)}
%real cases

\iffalse

other comments:

variation: 
garment type (done)
body shape (done)
background/illumination (done)

evaluation measures (L1,GAN,VGG,IOU,FID4vid) not sure:PSNR,SSIM,MSE (done)
different garment templates (done)
before/after post processing
Scale of the dataset

the number of sampled views (5 / 10 / 15) for every training frame
w/ and w/o background image
Normal discriminator vs temporal-and-spatial discriminator
skip connections vs. SPADE normalization
Supervision on the mask

\fi

\section{Conclusion and Future Work}
In this paper, we presented a novel neural rendering pipeline to jointly simulate and render dynamic garments. Our two-stage solution first generates intermediate 3D deformations of a coarse template, and then, by learning deep neural features attached to the coarse template, we synthesize the final appearance of the target garment. In order to capture dynamic appearance changes, we augment the learned neural features with motion features that encode the underlying body motion. Finally, our temporal-and-spatial discriminator ensures to produce plausible details in a temporally coherent manner. We evaluate our method thoroughly across unseen views, character body motion, environment illumination for various garment types and demonstrate the state of the art performance. We also test our framework on real video sequences to drive both the motion of the avatar and the draped garment dynamics.

\paragraph{Limitations and Future Work}
Our method has limitations, which we plan to address in future work. First, our learned network is character- and garment-specific. While we show how our network can be fine-tuned to handle different body shapes, a promising future direction is to employ few-shot learning techniques or investigate meta-learning to improve generalization. 

Second, we encode the garment appearance with the learned neural features. Hence, we need to learn a new set of neural features for different garment texture patterns. A possible solution is to train our rendering network to synthesize the UV coordinates and a shading layer instead. Such UV coordinates can then be used to sample different texture patterns. 

Third, powered by a CNN based network architecture and the patch-based generative method, our network can be trained with a small motion dataset with only 850 frames. The garment appearance generated for each frame is plausible, but some artifacts (e.g., temporal consistency and flickering) can be observed, especially when testing on a new motion that is not covered well by the distribution of training poses. Such artifacts can be largely resolved by enriching our training dataset with more motion clips and sampled views. 

Fourth, in our current setup, we do not model the garment intersection with body skeleton, or the interaction between the foreground character and the background scene. Therefore, our method does not generate realistic shadow. Possible solutions include modeling the shadow effects during training data generation or applying shadow prediction~\cite{sheng2020ssn} in post-processing. 

Finally, we presented a two-stage strategy and used a coarse template to bridge the feature domain and the RGB image domain. Jointly training the \textit{Joint2Coarse} and the rendering networks leads to an end-to-end system that may further enhance our performance. 

We show how our method can handle input motions predicted by human pose estimation methods. In the future, we would like to replace our background image rendering with a fully automatic image synthesis approach, e.g.,~\cite{chan2019everybody}, which provides satisfactory results when handling undressed 3D characters. This results in a pure neural network based system which can be used to retarget live performances of real actors to drive the synthesis of avatars wearing complex garments.

\bibliographystyle{ACM-Reference-Format}
\bibliography{main}
\end{document}